\documentclass[10pt,twocolumn,letterpaper]{article}

\usepackage{cvpr}              
\usepackage{cuted} 
\definecolor{cvprblue}{rgb}{0.21,0.49,0.74}
\usepackage[pagebackref,breaklinks,colorlinks,allcolors=cvprblue]{hyperref}
\usepackage{xcolor}
\usepackage{multirow}
\usepackage{makecell}
\usepackage{setspace}

\def\paperID{7248} 
\def\confName{CVPR}
\def\confYear{2026}

\definecolor{MyBlue}{HTML}{4472C4} 
\definecolor{MyOrange}{HTML}{ED7D31} 
\definecolor{MyGreen}{HTML}{70AD47} 
\definecolor{deepPink}{RGB}{255, 20, 147}

\title{{
{\color{MyBlue}Cycle}{\color{MyOrange}Manip}: Enabling Cyclic Task Manipulation \\ via Effective Historical Perception and Understanding
}}

\author{
    \textbf{Yi-Lin Wei}\textsuperscript{*,1},
    \, \textbf{Haoran Liao}\textsuperscript{*,1},
    \, \textbf{Yuhao Lin}\textsuperscript{1},
    \, \textbf{Pengyue Wang}\textsuperscript{1},
    \, \textbf{Zhizhao Liang}\textsuperscript{1},\\
    \textbf{Guiliang Liu}\textsuperscript{2,3},
    \, \textbf{Wei-Shi Zheng}\textsuperscript{†, 1,3} \\ 
    \textsuperscript{1} School of Computer Science and Engineering, Sun Yat-sen University \\
    ~\textsuperscript{2} The Chinese University of Hong Kong, Shenzhen
    ~\textsuperscript{3} Shenzhen Loop Area Institute \\
    \href{https://isee-laboratory.github.io/CycleManip/}{\textcolor{deepPink}{https://isee-laboratory.github.io/CycleManip/}}
    \vspace{-2em}
}

\begin{document}
\maketitle
\footnotetext[1]{Equal contribution.}
\footnotetext[2]{Corresponding author.}

\begin{strip}
\vspace*{-40pt}
  \centering
  \includegraphics[width=\textwidth]{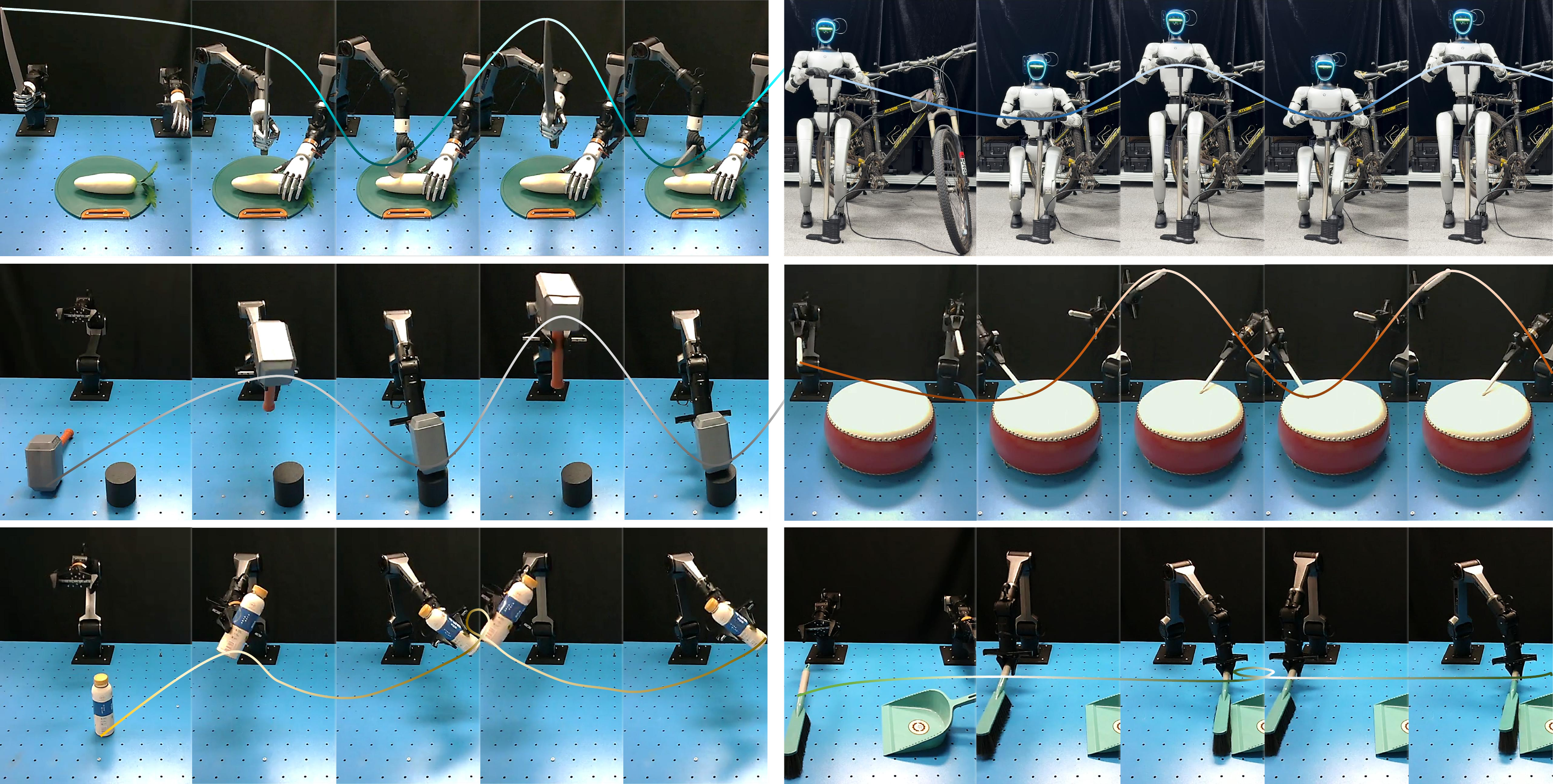}
  \captionsetup{skip=4pt}
  \captionof{figure}{Visualization of \textbf{CycleManip} performing various cycle-based manipulation tasks with different robot platforms.}
  \label{fig:teaser}
\vspace*{-5pt}
\end{strip}
\vspace{-0.5\baselineskip}

\begin{abstract}
\vspace{-2em}
\begin{spacing}{0.99}
In this paper, we explore an important yet underexplored task in robot manipulation: \textbf{cycle-based manipulation}, where robots need to perform cyclic or repetitive actions with an expected terminal time. These tasks are crucial in daily life, such as shaking a bottle or knocking a nail. However, few prior works have explored this task, leading to two main challenges: \textbf{1)} the imitation methods often fail to complete these tasks within the expected terminal time due to the ineffective utilization of history; \textbf{2)} the absence of a benchmark with sufficient data and automatic evaluation tools hinders development of effective solutions in this area. To address these challenges, we \textbf{first} propose the CycleManip framework to achieve cycle-based task manipulation in an end-to-end imitation manner without requiring any extra models, hierarchical structure or significant computational overhead. The core insight is to enhance effective history perception by a cost-aware sampling strategy and to improve historical understanding by multi-task learning. \textbf{Second}, we introduce a cycle-based task manipulation benchmark, which provides diverse cycle-based tasks, and an automatic evaluation method. Extensive experiments conducted in both simulation and real-world settings demonstrate that our method achieves high success rates in cycle-based task manipulation. The results further show strong adaptability performance in general manipulation, and the plug-and-play ability on imitation policies such as Vision-Language-Action (VLA) models. Moreover, the results show that our approach can be applied across diverse robotic platforms, including bi-arm grippers, dexterous hands, and humanoid robots.
\end{spacing}
\end{abstract}   
\vspace{-2em}
\section{Introduction}
\label{sec:intro}

\begin{figure*}[t!]
  \centering
  
  \begin{subfigure}{0.5\linewidth} 
    \includegraphics[width=\linewidth]{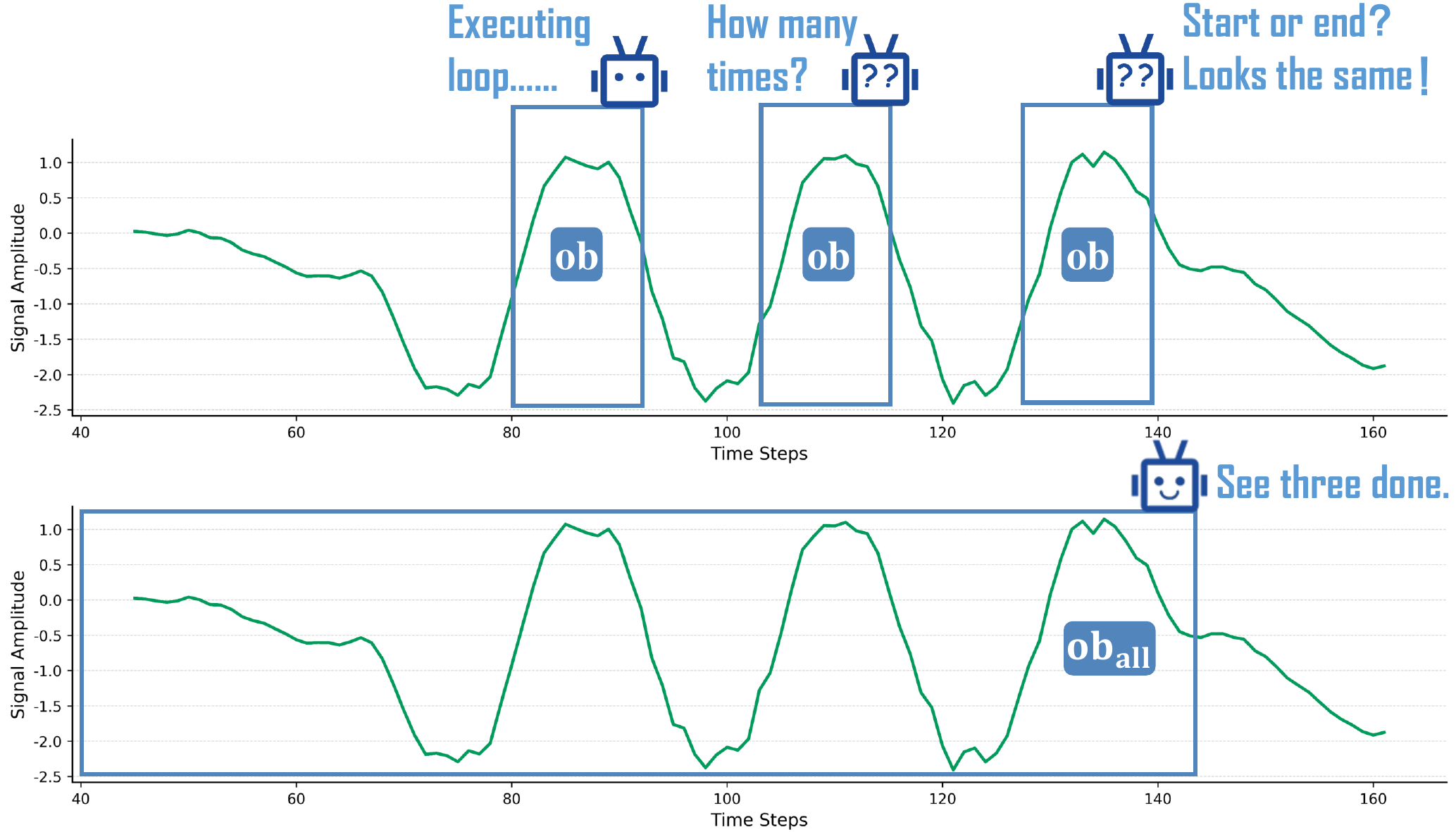}
    \caption{Necessity of Historical Perception}
    \label{fig:q1-zero}
  \end{subfigure}
  \begin{subfigure}{0.5\linewidth} 
    \includegraphics[width=\linewidth]{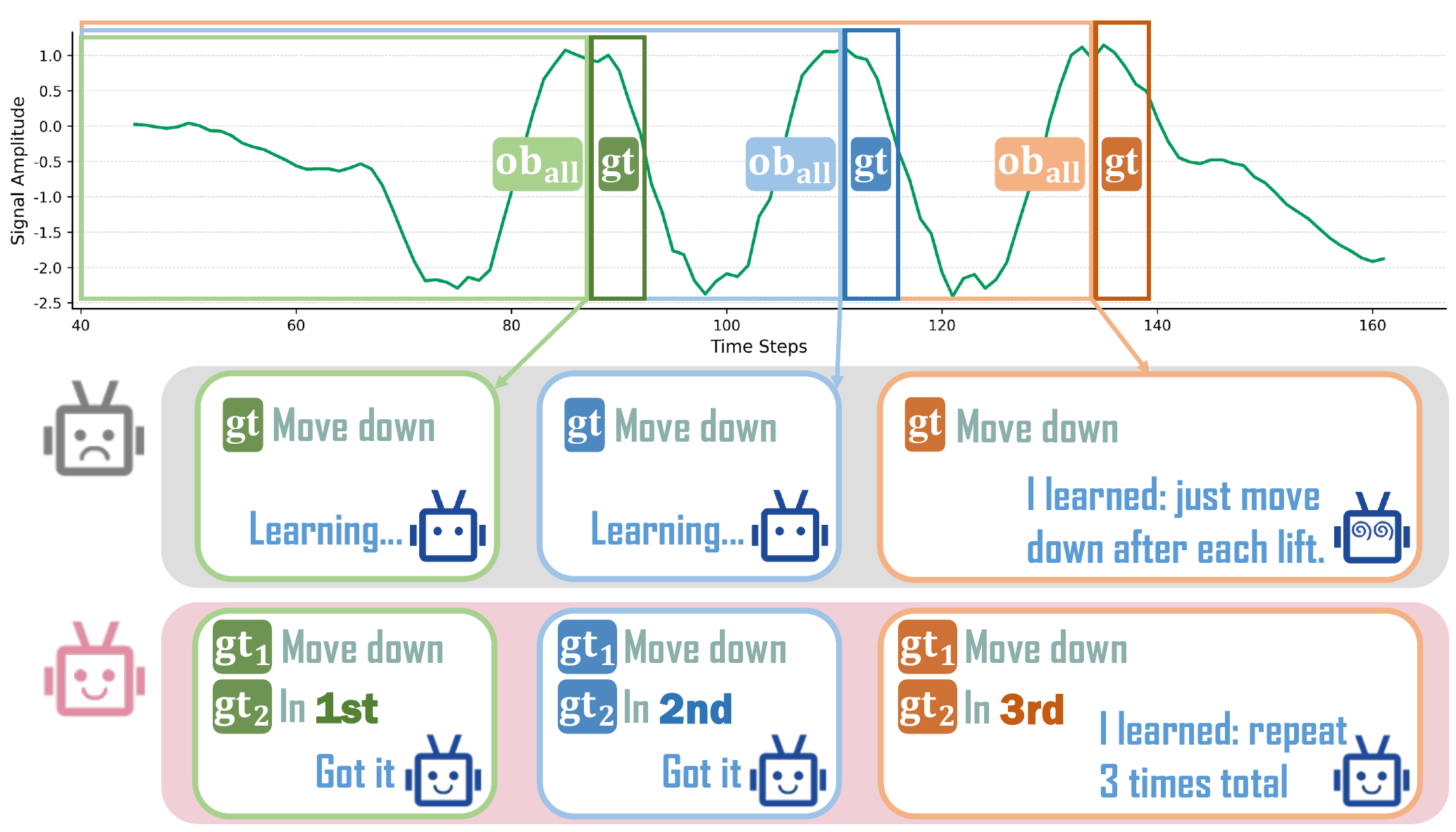}
    \caption{Necessity of Historical Understanding}
    \label{fig:q2-zero}
  \end{subfigure}
  
  \caption{Necessity of historical perception and understanding in cyclic manipulation. (a) The absence of historical perception, leaving the model unaware of the number of cycles executed in the past. (b) Relying solely on ground-truth imitation supervision produces identical targets across cycles, hindering the model’s sense of progression and reducing feature discriminability.
  }
  \label{fig: motivation}
\end{figure*}

The ability of a robot to autonomously take care of various tasks in our daily life is a long-term goal of the computer vision and robotics community~\cite{black2024pi0, zhao2023act, james2020rlbench, liu2023libero}. One key observation is that many tasks in the household involve repetitive and cyclic actions, such as dispensing several pumps of syrup or shaking a bottle until the contents are mixed~\cite{al2022impact,holland2020automation,rane2017assembly}. These tasks require the robot to perform cyclic actions and stop at an expected time, which presents significant challenges: the robot needs to execute repetitive actions and complete them within the desired time frame according to the user command or task execution status.  Previous work on robot manipulation has focused on general tasks~\cite{chi2025dp, liu2024rdt}; however, relatively little work has explored how to enable robots to perform cyclic tasks. Moreover, there is a lack of benchmarks that provide sufficient data and evaluation tools to assess the applicability of existing methods to cyclic tasks.

In this work, we explore this crucial task, CycleManip, the problem of robotic cyclic manipulation where a robot must accurately execute cyclic motions and stop at the correct moment. Such tasks are common in daily scenarios, but their inherently non-Markovian nature makes them challenging: the correct decision at any moment depends not only on the current observation, but also on the accumulated progress within the cycle. As a result, autonomous policies must model historical temporal dependencies and reason about the progression across multiple cycles\cite{whitehead1995reinforcement}. 

However, traditional imitation policies typically rely on short observation windows for action prediction~\cite{black2024pi0, ze20243dp3, brohan2022rt, kim2024openvla, team2024octo, wang2024rise, rdt2}, which leads to failure in cyclic manipulation tasks. This is because short observations across cycles often appear similar, causing the model to confuse its decisions as shown in Figure~\ref{fig: motivation} (a). For example, in a task that requires the robot to shake a bottle five times, the visual observation after each shaking step remains nearly identical, making it difficult for such models to infer whether the robot should continue or stop, since they cannot track how many shaking cycles have already been completed. An intuitive remedy is to expand the observation horizon, but doing so is costly: encoding and fusing high-dimensional visual observations at every timestep increases computation and latency. 

To empower robots with capabilities to complete cyclic manipulation tasks, we introduce the CycleManip framework within an end-to-end imitation manner, without requiring additional models, hierarchical structures, or significant computational overhead. The framework enables cyclic manipulation by enhancing historical perception and understanding. It consists of two core components: (1) effective historical perception: efficiently expanding the observation horizon without incurring substantial computational overhead, and (2) effective historical understanding: improving the policy’s ability to model cycle progression through multi-task learning. For the first component, we propose a cost-aware sampling strategy: sparse sampling for high-dimensional visual inputs to reduce overhead, and dense sampling for low-dimensional observations, such as proprioception, to capture temporal cycle characteristics. For the second component, we introduce a multi-task learning objective that encourages the policy to understand and infer the cycle stage. By jointly learning manipulation and cycle-stage prediction, the policy implicitly learns cycle features and makes better decisions on whether to continue or terminate the action, improving its reliability in cyclic manipulation.

To support our framework, we present a cycle-based manipulation benchmark that provides a diverse set of simulated tasks with automated data generation and model evaluation tools. The benchmark is built on the RoboTwin 2.0 simulation platform~\cite{chen2025robotwin2}, where we incorporate configurable cyclic action parameters into the data collection pipeline, enabling the generation of demonstrations with arbitrary numbers of repetitions and corresponding instructions. In addition, we develop an automated evaluation system in which an attempt is considered successful only if the policy not only completes the manipulation task but also performs the correct number of cycles.

Extensive experiments are conducted to validate the effectiveness of our framework in both simulation and on diverse real-world robotic platforms. The result shows that our framework significantly surpasses previous imitation methods. Furthermore, we demonstrate that our method also generalizes well to general manipulation tasks, and is also plug-and-play compatible with other imitation policies, such as Vision-Language-Action (VLA) models. Additionally, our approach is applicable across various robotic platforms, including bi-arm grippers, dexterous hands, and humanoid robots. In summary, our framework enables reliable cycle-based manipulation, representing a significant step toward more autonomous and adaptable robotic behavior in real-world environments.

\section{Related work}
\label{sec:formatting}
\subsection{Robotic Manipulation}
Robotic manipulation is a fundamental challenge in robotics and is considered an essential cornerstone for achieving Artificial General Intelligence (AGI). Current dominant methodologies, such as Imitation Learning (IL)~\cite{chi2025dp,ze20243dp3,zhao2023act} and Vision-Language-Action (VLA) models~\cite{zitkovich2023rt2,kim2024openvla,black2024pi0,intelligence2025pi05}, excel at modeling complex data distributions. They effectively predict subsequent actions based on current observations, demonstrating strong performance in sequential tasks. However, these approaches falter in cyclic tasks. When faced with similar inputs, the model struggles to distinguish the current phase of the cycle, which can lead to it becoming trapped in infinite loops or terminating prematurely. To solve this problem, we propose the CycleManip framework to enable cyclic task manipulation in an end-to-end imitation manner by the effective perception and understanding of historical information. 

\subsection{Historical Modeling}
Historical modeling is important for Large-Language-Models~\cite{wang2025MemoryLLM}, Vision-Language-Navigation~\cite{zhang2025MemoryVLN}, video generation~\cite{xiao2025worldmem} and robot manipulation~\cite{shi2025memoryvla}. The historical information is used to address the limitations of short-term observations, by retrieval~\cite{yu2025context,xiao2025worldmem}, large kernels~\cite{gu2025long}, memory caching~\cite{zeng2024poliformer, liu2025ttf}. Recent robot manipulation methods integrate historical visual information for visual memory~\cite{liu2025ttf,shi2025memoryvla,sridhar2025memer}, with a primary focus on long-horizon tasks. In contrast, this paper addresses the more challenging domain of cyclic tasks and introduces a cost-aware sampling strategy to integrate historical low-overhead proprioceptive information.

\subsection{Cyclic Task}
Cyclic tasks are important in daily life and represent a critical challenge that needs to be addressed for robot deployment. The execution of cyclic tasks can significantly enhance efficiency~\cite{hausser2014experimental, hall1988cyclic} and be crucial in environments such as factories, laboratories~\cite{al2022impact,holland2020automation,rane2017assembly} and human-robot collaboration scenarios~\cite{kheirabadi2023human, keshvarparast2024collaborative}. Some traditional works implement cyclic tasks by control strategies, which may be limited by poor adaptability to dynamic environments. Recently, some deep learning-based methods have been attempted for simple cyclic tasks~\cite{yang2022learning}; however, these methods are either limited to simple tasks with fixed cyclic times~\cite{torne2025learning}, limited to specific task scenarios~\cite{liang2020teaching,huang2019accurate}, or rely on external auxiliary models~\cite{chen2025history}. In this paper, we propose a method that enables diverse cyclic manipulation tasks with cyclic time control capability, in an end-to-end manner without leveraging any additional auxiliary models.

\section{CycleManip Framework}

\begin{figure*}
  \centering
  \includegraphics[width=\textwidth]{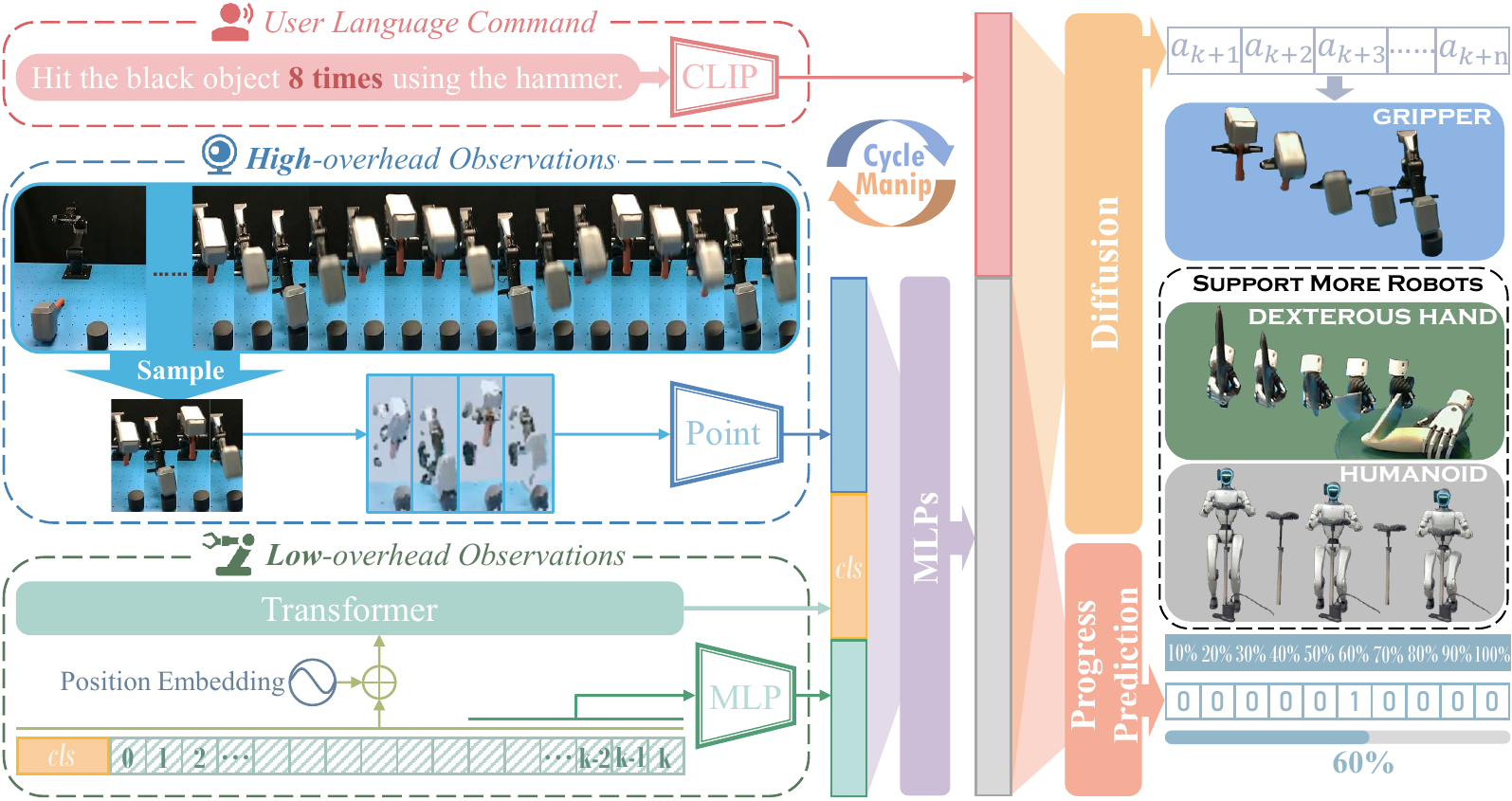}
  \caption{The overall framework. Given the user command and robot observation, the framework aims to execute operational tasks containing cyclic actions. We first employ cost-aware sampling strategy to achieve effective historical perception by different sampling for high and low overhead observation. Then all observation and language command are encoded as diffusion condition to predict robot action. Moreover, the observation features are employed to predict the task progress for better historical understanding.}
  \label{fig:methof}
\end{figure*}

\subsection{Overview}
In this paper, we explore the task of robot cyclic manipulation, where the robot is required to perform a cyclic manipulation action for a specified number of cycles based on a natural language instruction. For example, given the command “shake the bottle three times”, the robot must accurately grasp the object, execute shaking motion exactly three times, and autonomously terminate the action upon completion. 

To address this problem, we adopt an imitation learning paradigm to train a language-guided manipulation policy. Specifically, the policy accesses to the expert demonstration dataset, where each trajectory is represented as $\{lan,(o_1, a_1), (o_2, a_2), ..., (o_T, a_T)\}$, with $lan$ denote user language command, $o_t$ denoting the robot’s observation and $a_t$ the corresponding action at time $t$. The objective is to learn a policy $\pi$ that predicts the next action based on the historical observations, enabling the robot to perform and regulate cyclic actions in a closed-loop manner.
\begin{equation}
a_t = \pi(lan,\{o_i\}_{i=1}^{t}).
\end{equation}

\subsection{Effective Historical Perception}
\textbf{Challenge of Historical Perception.} The cyclic task, as a non-Markovian process, demands the policy to rely not only on the current observation but also on historical information, as shown in Figure \ref{fig: motivation} (a). For instance, in the task of'shake a bottle five times', the current observation remains similar after each shake, making it difficult for the model to determine whether to continue shaking or stop, as it does not know how many more shakes are needed or if the task is completed. An intuitive solution is to expand the model's observation horizon, however this approach introduces substantial computational overhead, since encoding and fusing high-dimensional visual observations at every timestep is expensive. 


\textbf{Effective Historical Perception.} To address these challenges, we propose a cost-aware history sampling strategy that employs different sampling methods for different information. Specifically, we first categorize the observations into two types: low-overhead observations $o^{l}_i$ (e.g., robot proprioception) and high-overhead observations $o^{h}_i$ (e.g., RGB images or point clouds). For low-overhead observations, we employ a dense and broad sampling strategy $\mathcal{H}_l$, which maintains low computational cost due to the inexpensive encoding of such observations. For high-overhead observations, we employ a heuristic frame sampling strategy $\mathcal{H}_h$ to sample visual observations with greater diversity, avoiding an increase in the number of samples. Consequently, the policy is formulated as:
\begin{equation}
a_t = \pi\left( \mathcal{H}_h \left(\{{o^{h}_{i}}\}_{i=1}^{t} \right), \mathcal{H}_l \left(\{{o^{l}_{i}}\}_{i=1}^{t} \right) \right).
\end{equation}

To construct a low-overhead observation that is both compact and representative of the cyclic manipulation process, we use the pose difference of the end effector. There are two key points: 1) The cyclic nature of the end effector is more apparent and easier to model compared to joint positions, as it directly reflects the overall movement pattern; 2) Using pose differences helps mitigate the bias introduced by absolute positions, enabling the model to focus more on the cyclic nature of the task itself. This compact and representative observation allows us to extend the temporal observation range while keeping the computational cost low. In our experiments, we incorporate all past low-overhead observations into the sampling process.

For high-overhead observations (e.g., point clouds for 3D-based methods and RGB images for 2D-based methods), we adopt a heuristic sampling strategy that selects past frames with a longer observation horizon while keeping the overall number of frames $K_{\text{high}}$ consistent with the original version. Specifically, given the first frame as $0$ and the current frame as $t$, we perform right-sided binary sampling to collect $0.5 \cdot K_{\text{high}}$ frames. And we apply exponential sampling from the latest frame $t$, collecting another $0.5 \cdot K_{\text{high}}$ frames according to the rule $t - 2^k$, where $k$ denotes the sampling index.

\begin{figure*}
  \centering
  \includegraphics[width=\textwidth]{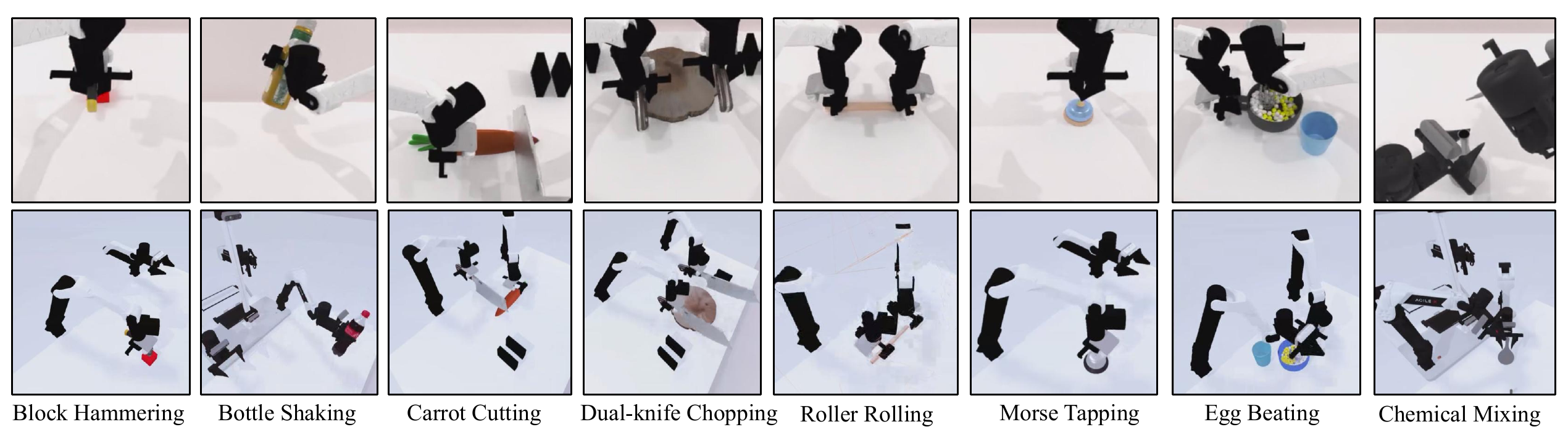}
  \vspace{-2em}
  \caption{The visualization of the tasks in CycleManip Benchmark.}
  \label{fig:benchmark}
  \vspace{-1em}
\end{figure*}

\subsection{Effective Historical Understanding}
\textbf{Challenge of Historical Understanding.}
While expanding the observation horizon provides the model with broader perception, an overload of information introduces challenges in feature encoding and understanding. Relying only on imitation learning supervision typically fails to enable the model to truly understand the inherent temporal process of cyclic tasks, as shown in Figure \ref{fig: motivation} (b). For example, in a hammering task, the historical observations before each strike differ due to different motion histories, yet the supervision signal remains the same (conduct a hammering). This discrepancy may force the model to learn features that converage to a local optimum, making it difficult for the model to capture the temporal information necessary for cyclic behavior.

\textbf{Effective Historical Understanding.}
To address this challenge, we employ a multi-task learning strategy to encourage the model to learn progression-discriminative features. Specifically, we introduce an auxiliary task that predicts the current phase of the overall process (e.g., the current cycle count or task progress). This encourages the model to learn distinct feature representations for different stages of the cycle, as the supervisory signals change during task progression. It is worth noting that, to mitigate the risk of overfitting the multi-task head, we first apply a multi-layer MLPs for feature fusion (which contributes to subsequent decisions in the diffusion model), followed by a single-layer MLP to predict the current progress $b_t$. The ground truth of the current progress is obtained by dividing the current frame number by the maximum frame number of this task. Then we uniformly partition the interval $[0,1]$ into ten bins and discretize $b_t$ into the corresponding class label $y_t$, which is then used for a 10-way classification objective.

\subsection{Framework Architecture}
Given the user instruction $lan$ and historical observation $\{o_i\}_{i=1}^{t}=\{o_{high}, o_{low}\}_{i=1}^{t}$, we first employ a cost-aware sampling strategy to obtain sampled observation $\mathcal{H}_h \left(\{{o^{h}_{i}}\}_{i=1}^{t} \right)$ and $\mathcal{H}_l \left(\{{o^{l}_{i}}\}_{i=1}^{t} \right)$. The language feature $f_{lan}$ is encoded by CLIP encoder~\cite{radford2021clip}, the high observation feature $f_{h}$ is encoded by a point encoder~\cite{ze20243dp3}. For low-overhead observations, we employ a Transformer encoder~\cite{vaswani2017attention} to obtain $f_{l}$, where the global features are derived from the CLS token, and local features are extracted using an MLP from the recent frames. Then we employ MLPs to fuse $f_{l}$ and $f_{h}$ obtain $f_{lh}$, which will be used for diffusion decision and auxiliary task prediction. Finally, we concatenate the language feature $f_{lh}$ and observation feature $f_{lh}$ as the condition feature for diffusion model. We employ film conditioning \cite{peebles2023dit} to output the action prediction. The final loss function of the model is:
\begin{equation}
    \mathcal{L} = \alpha *\text{MSE}(a_t, a_t^*) + \beta * \text{CE}(b_t, b_t^*),
\end{equation}
where $a_t^*$ and $b_t^*$ are the ground truths of the action and auxiliary task target. We employ Mean Squared Error (MSE) and Cross-Entropy (CE) losses.


\section{CycleManip Benchmark}
\label{sec: benchmark}

To support our framework, we build a benchmark for cyclic manipulation tasks based on the RoboTwin 2.0 platform \cite{chen2025robotwin2}. We collect 8 cyclic manipulation task environments for convenient data collection and policy evaluation, as shown in Figure \ref{fig:benchmark}. More details can be found in the supplementary material.

In data collection, we integrate loop control functionality into the data collection pipeline of RoboTwin. Specifically, we predefine the initiation and termination points of a single cycle of each task, followed by iterative repetition of this process according to the desired number of cycles. We collect 200 expert demonstration trajectories for each task with loop times ranging from 1 to 8, where each demonstration is annotated with the target cycle time, the completed number of cycles each time step, and the object's 6D pose. Further dataset details can be found in the supplementary materials.  

\begin{table*}[t]
\centering
\setlength{\tabcolsep}{3pt} 
\renewcommand{\arraystretch}{1.2} 
\begin{tabular}{l|llllllllllllllll}
\toprule
     & \multicolumn{2}{l}{\makecell{block \\ hammering}} 
     & \multicolumn{2}{l}{\makecell{bottle \\ shaking}} 
     & \multicolumn{2}{l}{\makecell{roller \\ rolling}} 
     & \multicolumn{2}{l}{\makecell{carrot \\ cutting}} 
     & \multicolumn{2}{l}{\makecell{dual-knife \\ chopping}} 
     & \multicolumn{2}{l}{\makecell{egg \\ beating}} 
     & \multicolumn{2}{l}{\makecell{chemical \\ mixing}} 
     & \multicolumn{2}{l}{\makecell{morse \\ tapping}} \\ \cline{2-17}
     & $Suc.$ & $Cyc.$ & $Suc.$ & $Cyc.$ & $Suc.$ & $Cyc.$ & $Suc.$ & $Cyc.$ & $Suc.$ & $Cyc.$ & $Suc.$ & $Cyc.$ & $Suc.$ & $Cyc.$ & $Suc.$ & $Cyc.$  \\ \midrule
DP   &8   &8.33  &8    &7.91  &25   &1.88  &4   &5.65  &  8  & 3.79  &  15  & 2.18  & 20   & 1.16  & 0   &  - \\
DP3  &23 & 5.55 &16  & 4.58 &33 & 1.44 &38 & 1.92 &48  & 0.81  & 19   & 1.95  & 18  &  1.41  & 1   & -  \\
RDT  &20 & 2.15 &15  & 1.53 &35 & 1.55 &36 & 1.24 &42  & 2.13  & 16 & 2.31 & 12 &  2 & 0  &  - \\
Pi-0  &13 & 3.44 &19  & 2.00 &14 & 3.80 &8  & 2.54 &1   & 3.14  & 4  & 2.15 & 2  & 2.37  & 0  &  - \\ 
\midrule
Ours &86 & 0.25 &95  & 0.29 &97 & 0.03 &86 & 0.81 &90  & 0.4  & 74  & 0.61  & 53  & 0.76  & 91  & -  \\ \bottomrule
\end{tabular}
\caption{\textbf{Performance comparison on various cyclic manipulation tasks.} Our method outperforms all baselines in Success Rate (Suc. \%) and Cycle Count Deviation (Cyc.). Since morse tapping has a fixed cycle count, we do not report its Cyc.}
\label{tab: simulation}
\vspace{-1em}
\end{table*}

In evaluation, we design an automatic cycle evaluation system to determine whether the model successfully completes the cyclic task for the expected number of iterations. We achieve this by analyzing the most distinctive characteristics of cyclic motion of each task, such as the object poses in the bottle-shaking task and the contact signal between the hammer and block in the block-hammering task. Specifically, for tasks involving physical contact such as hammering and cutting, we use a state-machine-based collision detection system to count the number of successful cycles. For non-contact tasks such as shaking and stirring, the peak detection algorithm is used to estimate the cycle times based on the object pose trajectory. After each evaluation episode, a detailed loop detection report will be generated, including the total number of cycles completed, the time step of each cycle, and whether the task was successfully completed. To ensure its accuracy and reliability, the system has been rigorously tested through manual review. For each task, we manually checked 100 results and confirmed that the automatic evaluation system is reliable.

\section{Experiment}
\subsection{Experimental Setup}

\textbf{Real-world Setup}.
We conduct real-world experiments across diverse and heterogeneous robotic embodiments, including single-arm and dual-arm grippers (AgileX Piper), dexterous hands (BrainCO Revo2), and a humanoid platform (Untree G1). The visual observations are captured using an Intel RealSense L515 depth camera. There are six real-world cyclic manipulation tasks: block hammering, bottle shaking, drum beating, tire pumping, knife cutting, table cleaning. For each task, 50–150 teleoperated demonstrations are collected using Gello~\cite{wu2024gello} for gripper arms, TypeTele~\cite{lin2025typetelereleasingdexterityteleoperation} for dexterous hands, and OpenWBC~\cite{liu2025openwbc, ben2025homie} for the humanoid platform as shown in Figure \ref{fig:hardware_setup}. Additional details are provided in the supplementary material.

\textbf{Simulation Setup}
We further evaluate our approach in simulation using the CycleManip benchmark and the RoboTwin 2.0 benchmark. In CycleManip simulation benchmark, two ARX-X5 robotic arms and a RealSense D435 head camera are employed, and eight cyclic tasks are performed as described in Section \ref{sec: benchmark}. For RoboTwin 2.0 benchmark, we follow the same experimental settings as in \cite{RoboTwin2025Leaderboard, chen2025robotwin2} and evaluate four general manipulation tasks. More details are provided in the supplementary material.

\textbf{Evaluation Metrics}.
We use two metrics for evaluating cyclic manipulation performance: 1) Success Rate ($Suc.$), which measures the proportion of trials in which the task is successfully completed and the required number of cycles is achieved; 2) Cycle Count Deviation ($Cyc.$), which quantifies the average absolute deviation between the executed and ground-truth cycle counts. In addition, for general tasks, we adopt the Success Rate metric following the RoboTwin 2.0 benchmark protocol. All simulation experiments are conducted over 100 trials, while real-world experiments were evaluated over 16 trials.

\subsection{Implementation Details}
\textbf{Training and Inference Details}
For our framework, the number of sampled high-overhead observations $K_{high}=6$ and the action horizon is 8. We employ DDIM \cite{song2020ddim} as diffusion sampler with 100 training steps and 10 testing steps. For training, the number of epochs is set to 300 and the batch size is 128. The loss weights $\alpha=1$ and $\beta=0.1$. The initial learning rate is \( 2.0 \times 10^{-4} \) with a cosine learning rate scheduler \cite{loshchilov2016sgdr}. The experiments of our framework are conducted using PyTorch on a single RTX 4090 GPU. All real world experiments are inferred in a single RTX 4070 GPU. More details are provided in the supplementary material.

\textbf{Baseline Reproduction}
We reproduce all baselines using the default settings of RoboTwin across all experiments. For the DP \cite{chi2025dp} and DP3 \cite{ze20243dp3} baselines, since they do not include a language encoder, we implement an identical language encoder within our framework for fair comparison. The Pi-0 \cite{black2024pi0} and RDT \cite{liu2024rdt} models are trained on a single H100, while DP and DP3 models are trained on a single RTX 4090. More details are provided in the supplementary material.

\begin{figure*}
  \centering
  \includegraphics[width=0.8\textwidth]{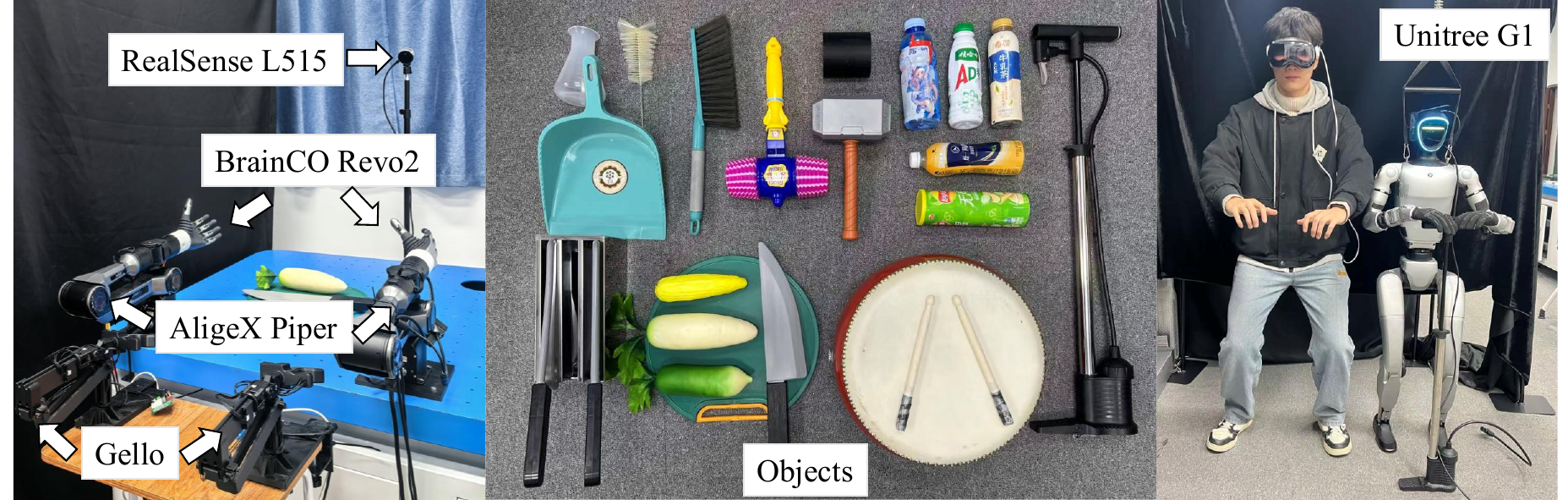}
  \caption{\textbf{The hardware setup for real-world benchmark.} (a) AgileX Piper robot (gripper and BrainCO Revo2 dexterous hands) used for single-arm and dual-arm tasks, equipped with an Intel RealSense L515 RGB-D camera for visual perception. (b) Object sets for real-world tasks.
   (c) Unitree G1 humanoid robot utilized for whole-body cyclic tasks.}
  \label{fig:hardware_setup}
\end{figure*}

\begin{table*}[]
\centering
\begin{tabular}{ll|llllllll}
\toprule
\multicolumn{1}{c}{\multirow{2}{*}{task}} & \multicolumn{1}{c|}{\multirow{2}{*}{setting}} 
& \multicolumn{2}{c}{DP3~\cite{ze20243dp3}} 
& \multicolumn{2}{c}{w/o Task} 
& \multicolumn{2}{c}{Ours} \\ \cline{3-8} 
\multicolumn{1}{c}{} & \multicolumn{1}{c|}{} 
& \multicolumn{1}{c}{$Suc.$} & \multicolumn{1}{c}{$Cyc.$} 
& \multicolumn{1}{c}{$Suc.$} & \multicolumn{1}{c}{$Cyc.$}  
& \multicolumn{1}{c}{$Suc.$} & \multicolumn{1}{c}{$Cyc.$} \\ \midrule
block hammering       & Single Gripper   & 37.5  & 1.12  & 62.5  &0.5   & 93.75  & 0.125   \\
bottle shaking        & Single Gripper   & 12.5   &3.81   &31.25   &1.31 & 68.75  & 0.375  \\
drum beating          & Bi-Gripper       & 0  & 2.4  & 60  & 0.8  & 90  &0.2   \\
table cleaning        & Bi-Dexterous     & 20 & 0.9 & 40  & 1.6  &  100  & 0.00  \\
tire pumping          & Humanoid         & 10  &  3.70 &  20 &  2.0 &  50 & 1.5  \\
knife cutting         & Bi-Dexterous     & 0 & 1.75  & 25  & 4.125  &  75  & 0.88   \\
 
\bottomrule
\end{tabular}
\caption{Results of Real-World Experiments. (w/o Task means Ours without historical understanding.) }
\label{table: Real world}
\end{table*}

\subsection{Comparison with State-of-the-Art Methods}
We conduct a comprehensive evaluation of our proposed method against several state-of-the-art (SOTA) baselines, including DP, DP3, RDT, and Pi-0, with performance comparisons detailed in Table~\ref{tab: simulation}. To ensure a fair comparison, all baseline methods were reproduced and evaluated under the same experimental settings, following the protocol of the RoboTwin 2.0 benchmark. The results demonstrate the superior performance of our approach across both primary evaluation metrics. The failure of SOTA baselines stems primarily from their reliance on short observation windows. Pi-0 exemplifies this issue, predicting actions using only the current 1-frame observation, which results in the lowest overall success rates. Other baselines use slightly longer temporal windows, performing marginally better but still failing to reliably model the cyclic progression. Beyond task success, our average cycle deviation ($Cyc.$) is lower than baselines significantly, proving our method's superior perception of task progression. Overall, these results confirm our CycleManip framework, by enhancing historical perception and understanding, solves core cyclic manipulation challenges that hinder SOTA methods.

\begin{table*}[]
\centering
\begin{tabular}{l|ccccccc}
\toprule
     & \multicolumn{1}{l}{\makecell{place cans \\ plasticbox}} & \multicolumn{1}{l}{\makecell{Handover \\ Block}} & \multicolumn{1}{l}{\makecell{Pick Diverse \\Bottles}} & \multicolumn{1}{l}{\makecell{Stamp\\ Seal}} & \multicolumn{1}{l}{\makecell{Place Bread \\Basket}} & \multicolumn{1}{l}{\makecell{Open \\Microwave}} & \multicolumn{1}{l}{\makecell{Turn \\Switch}} \\ \midrule
RDT \cite{liu2024rdt}  & 6                                           & 45                                 & 2                                        & 1                              & 10                                     & 37                                 & 35                              \\
Pi-0 \cite{black2024pi0} & 34                                          & 45                                 & 27                                       & 3                              & 17                                     & 80                                 & 27                              \\
DP \cite{chi2025dp}   & 40                                          & 10                                 & 6                                        & 2                              & 14                                     & 5                                  & 36                              \\
DP3 \cite{ze20243dp3}  & 48                                          & 70                                 & 52                                       & 18                             & 26                                     & 61                                 & 46                              \\ \midrule
Ours & 91                                          & 96                                 & 84                                       & 38                             & 61                                     & 93                                 & 64                              \\ \bottomrule
\end{tabular}
\caption{The comparison results in general manipulation from RoboTwin 2.0 Benchmark~\cite{RoboTwin2025Leaderboard}. Our method not only yields benefits for cyclic tasks but also achieves good performance for general manipulation tasks.}
\vspace{-1em}
\label{table:General Manipulation}
\end{table*}

\begin{table}[]
\begin{tabular}{l|cccc}
\toprule
       & {\makecell{bottle \\ shaking}} & {\makecell{rooller \\ rolling}} & {\makecell{carrot \\ cutting}}& {\makecell{dual-knife \\ chopping}} \\ \midrule
Pi-0\cite{black2024pi0}   &19   &14   &8  & 1  \\
Pi-0 + Ours   &72   & 69  &47 &41   \\ \midrule
\end{tabular}
\caption{Plug and Play Experiments. Our method cam serve as a plug-and-play component integrated into VLA imitation learning models, yielding significant performance improvements.}
\label{table: Plug and Play}
\end{table}

\begin{table}[]
\begin{tabular}{l|ccccc}
\toprule
     & $Suc.$ & $\text{Time}_\text{train}$   & $\text{GPU}_\text{train}$                 & $\text{Time}_\text{test}$      & $\text{GPU}_\text{test}$ \\ \midrule
DP3  & 38  & 0.073 & 16796 & 0.0893 & 5801                         \\
Ours & 86    & 0.102 &   17342   & 0.0953 & 6003          \\ \bottomrule
\end{tabular}
\caption{Efficiency analysis of our framework. Our method enhances historical awareness capability without significantly increasing computational overhead.}
\vspace{-1em}
\label{table: Efficiency analysis}
\end{table}

\textbf{Real-world Experiments.} To further validate the robustness and practical applicability of our framework, we conducted real-world experiments, with results presented in Table~\ref{table: Real world}. For these evaluations, we selected DP3 as the baseline, as it demonstrated the strongest performance among prior methods in our simulation benchmarks. For fairness, both policies were trained on the same demonstration data and tested under an identical physical setup. The results demonstrate that our framework outperforms baseline methods significantly, which is consistent with the findings from simulation experiments. Consequently, real-world results further validate the effectiveness and reliability of our proposed framework for practical cyclic manipulation tasks.

\subsection{Effectiveness of Historical Perception and Understanding}
To evaluate the effectiveness of our framework's core components, we conducted ablation studies on historical perception and understanding components, as shown in Table~\ref{table: Real world}. \textbf{(1) Historical Perception is effective.} By augmenting the baseline with our cost-aware sampling strategy module (w/o Task), we observe a significant boost in performance across all tasks. This demonstrates that our sampling strategy provides the policy with comprehensive historical observations, enabling it to accurately assess progress within the cyclic task. \textbf{(2) Historical Understanding is effective.} The subsequent integration of Historical Understanding (Ours) leads to a further improvement in performance. The results indicate that our multi-task objective shifts the model’s learning from simply perceiving history to actively understanding its structure for the current decision. By explicitly predicting its current progress, the policy develops a more discriminative feature space that enhances its awareness of the task's stage. This enhanced understanding is what ultimately enables the model to achieve highly reliable and precise execution of cyclic tasks.

\subsection{Results in general manipulation}
Table~\ref{table:General Manipulation} presents a comprehensive comparison of various methods in general manipulation tasks from RoboTwin 2.0 benchmark~\cite{chen2025robotwin2}. Our method achieves the best performance across all tasks, far surpassing baselines such as RDT, Pi0, DP, and DP3. This consistent superiority demonstrates our approach’s remarkable robustness, generalization capability, and effectiveness in handling diverse robotic manipulation scenarios, validating its potential for advancing the state-of-the-art in this field.

\subsection{Results on Heterogeneous Embodiments}
The real-world experiments are conducted on diverse robotic platforms, as shown in Figure \ref{fig:teaser}. For each platform, we extend the proprioceptive sensing and action dimensions to match the robot’s specific configurations, followed by model retraining using data collected from that particular robot. As evidenced by Table \ref{table: Real world}, our model exhibits strong adaptability to heterogeneous embodiments—including single-gripper, bi-gripper, humanoid, and bi-dexterous robots—and delivers robust performance across all these diverse robotic forms.

\subsection{Plug and Play Experiment}
Table~\ref{table: Plug and Play} demonstrates the effectiveness of plugging our method into other imitation policies, such as Vision-Language-Action model, Pi-0~\cite{black2024pi0}. In implementation, we only sample the current frame for visual observations, then encode all past joints via a transformer before feeding them into the action expert of Pi-0. The results show that our method significantly improves the performance, highlighting its robustness and adaptability.

\subsection{Efficiency Analysis}
Table~\ref{table: Efficiency analysis} compares the efficiency of our method with that of the baseline~\cite{chi2025dp} on the carrot-cutting task in our benchmark using a single RTX 4090 GPU.  $\text{Time}_\text{train}$ denotes the per-step training time (1 diffusion step), and $\text{Time}_\text{test}$ denotes the inference time (10 diffusion steps). $\text{GPU}_\text{train}$ and $\text{GPU}_\text{test}$ indicate GPU memory usage. Our method achieves a higher success rate ($Suc.$) while incurring only marginally increased training and inference time, as well as slightly higher GPU memory consumption.

\section{Conclusion}
We believe that enabling cyclic manipulation is a critical step toward autonomous robotic behavior in real-world scenarios. In this paper, we tackle the core challenges of cyclic manipulation, ineffective historical perception and understanding, by proposing the CycleManip framework. Our key insight is to enhance both historical perception via a cost-aware sampling strategy and historical understanding through multi-task learning, in an end-to-end imitation manner without extra modules and heavy computational overhead. To support our framework, we introduce a benchmark with diverse cyclic tasks, automated data generation, and evaluation tools. Extensive experiments in simulation and real-world settings validate that our method outperforms SOTA baselines across cyclic tasks and general manipulation tasks. Moreover, our framework supports plug-and-play integration with VLA models, and can adapt to heterogeneous robotic embodiments. In summary, this work explores enabling robots to reliably perform a broader range of daily tasks.

\section{Acknowledgment}
This work was supported partially by NSFC (92470202), Guangdong NSF Project (No. 2023B1515040025), Guangdong Key Research and Development Program (No. 2024B0101040004, No. 2025B0909020002). We would like to thank Jiangran Lyu for insightful and helpful discussions, and Pengxiang Liu for his work on robot hardware.

{
    \small
    \bibliographystyle{ieeenat_fullname}
    \bibliography{main}
}
\clearpage
\end{document}